%% file: main.tex
\documentclass[runningheads]{llncs}

\usepackage{graphicx}

\usepackage[disable]{todonotes}
\usepackage{amsmath}
\usepackage{amssymb}

\usepackage{endnotes} \let\footnote=\endnote 
\usepackage{subcaption}
\usepackage{hyperref}

\usepackage{algorithm}
\usepackage{algorithmic}

\usepackage{xspace}

\usepackage{booktabs}
\usepackage{tabularx}
\usepackage{multirow}
\usepackage{makecell}

\usepackage[utf8]{inputenc}

\newcommand{\ed}[1]{#1}

\newcommand{\approach}{\textsc{NCA}\xspace}

\usepackage{gensymb}

\title{Towards Neural Schema Alignment for OpenStreetMap and Knowledge Graphs}

\titlerunning{Towards Neural Schema Alignment for OSM and KGs}

\author{
Alishiba Dsouza$^{1}$
\href{https://orcid.org/0000-0001-5884-6234}{\includegraphics[scale=0.125]{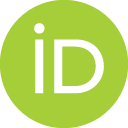}}
, 
Nicolas Tempelmeier$^{2}$
\href{https://orcid.org/0000-0003-0911-6264}{\includegraphics[scale=0.125]{Images/ORCIDiD_icon128x128.png}}
, 
Elena Demidova$^{1}$
\href{https://orcid.org/0000-0002-5134-9072}{\includegraphics[scale=0.125]{Images/ORCIDiD_icon128x128.png}}
}

\institute{
$^1$Data Science \& Intelligent Systems (DSIS), University of Bonn, Bonn, Germany \\
$^2$L3S Research Center, Leibniz Universit\"at Hannover, Hannover, Germany \\
  \email{dsouza@cs.uni-bonn.de} \email{tempelmeier@L3S.de}\\ \email{elena.demidova@cs.uni-bonn.de}
} 
 
\authorrunning{Alishiba Dsouza, Nicolas Tempelmeier and Elena Demidova}
 
\begin{document}

\maketitle

\begin{abstract}
OpenStreetMap (OSM) is one of the richest openly available sources of volunteered geographic information. Although OSM includes various geographical entities, their descriptions are highly heterogeneous, incomplete, and do not follow any well-defined ontology. Knowledge graphs can potentially provide valuable semantic information to enrich OSM entities. However, interlinking OSM entities with knowledge graphs is inherently difficult due to the large, heterogeneous, ambiguous and flat OSM schema and the annotation sparsity.  
This paper tackles the alignment of OSM tags with the corresponding knowledge graph classes holistically by jointly considering the schema and instance layers. 
We propose a novel neural architecture that capitalizes upon a shared latent space for tag-to-class alignment created using linked entities in OSM and knowledge graphs.
Our experiments performed to align OSM datasets for several countries with two of the most prominent openly available knowledge graphs, namely, Wikidata and DBpedia, demonstrate that the proposed approach outperforms the state-of-the-art schema alignment baselines by up to 53 percentage points in terms of F1-score. 
The resulting alignment facilitates new semantic annotations for over 10 million OSM entities worldwide, which is more than a 400\% increase compared to the existing semantic annotations in OSM.
\end{abstract}

\keywords{OpenStreetMap, Knowledge Graph, Neural Schema Alignment}

\input{01_intro}

\input{02_problem}
\input{03_approach}
\input{04_setup}

\input{05_evaluation}

\input{06_related_work}

\input{07_conclusion}

\subsubsection*{Acknowledgements} This work was partially funded by DFG, German Research Foundation (``WorldKG'', 424985896, DE 2299/2-1), BMBF, Germany (``Simple-ML'', 01IS18054) and BMWi, Germany (``d-E-mand'', 01ME19009B).

\bibliographystyle{splncs04}
\bibliography{references}

\footnotesize{
\theendnotes
}
\end{document}

%% file: 01_intro.tex
\section{Introduction}
\label{sec:introduction}

OpenStreetMap (OSM) has evolved as a critical source of openly available geographic information globally, including rich data from 188 countries. 
This information is contributed by a large community, which currently counts more than 1.5 million volunteers. 
OSM captures a vast, and continuously growing number of geographic entities, currently counting more than 6.8 billion \cite{osmstats}. 
The descriptions of OSM entities consist of heterogeneous key-value pairs, so-called \emph{tags}, and currently include over 80 thousand distinct keys. 
As OSM keys and tags do not possess any machine-readable semantics, OSM data is not directly accessible for semantic applications.

Whereas knowledge graphs (KGs) can provide precise semantics for geographic entities, large publicly available general-purpose knowledge graphs like Wikidata \cite{articlewikidata}, DBpedia \cite{articledbpedia}, YAGO \cite{10.1007/978-3-030-49461-2_34}, and even specialized KGs like EventKG \cite{GottschalkD19} and LinkedGeoData \cite{10.5555/2590208.2590210} lack coverage of geographic entities.
For instance, in April 2021, 126,894 entities with the tag \texttt{highway=road} were present in OSM, whereas Wikidata included only 66,114 entities for the equivalent class ``road''.
We believe that an alignment of OSM and knowledge graphs at the schema level can make a wide variety of geographic entities in OSM accessible through semantic technologies and applications.

The problem of schema alignment between OSM and knowledge graph is particularly challenging due to several factors, most prominently including the heterogeneous representations of types and properties of geographic entities via OSM tags, unclear tag semantics,
the large scale and flatness of the OSM schema, and the sparseness of the existing links.
OSM does not limit the usage of keys and tags by any strict schema and provides only a set of guidelines\footnote{OSM ``How to map a": \url{https://wiki.openstreetmap.org/wiki/How_to_map_a}}. 
As a result, the types and properties of OSM entities are represented via a variety of tags that do not possess precise semantics.  
As an example, consider an excerpt from the representations of the entity 
``Zugspitze" (the highest mountain in Germany) in Wikidata and OSM:\\
%
\input{example}

\noindent In Wikidata, an entity type is typically represented using the \texttt{instance of} property. In this example, the statement ``Q3375 \texttt{instance of} mountain" indicates the type ``mountain" of the entity ``Q3375".
In OpenStreetMap, the type ``mountain" of the same entity is indicated by the tag \texttt{natural=peak}. 
As OSM lacks a counterpart of the \texttt{instance of} property, it is not clear which particular tag represents an entity type and which tags refer to other properties. Furthermore, multiple OSM tags can refer to the same semantic concept.
Finally, whereas the OSM schema with over 80 thousand distinct keys is very extensive, 
the alignment between OSM and knowledge graphs at the schema level is almost nonexistent. For instance, as of April 2021, Wikidata contained 585 alignments between its properties and OSM keys, which corresponds to only 0.7\% of the distinct OSM keys. 
Overall, the flatness, heterogeneity, ambiguity, and the large scale of OSM schema, along with a lack of links, make the alignment particularly challenging.

Existing approaches for schema alignment operate at the schema and instance level and consider the similarity of schema elements, structural similarity, and instance similarity. 
As OSM schema is flat, ontology alignment methods that utilize hierarchical structures, such as \cite{NgoBT13,madhavan2001generic}, are not applicable.
A transformation of OSM data into a tabular or relational format leads to highly sparse tables with a large number of properties. 
Therefore, approaches to syntactic or instance-based alignment for relational or tabular data, such as, e.g., \cite{ZhangB20,DemidovaON13}, or syntactic matching of schema element names \cite{inproceedingsunal} cannot yield good results for matching OSM tags with KG classes.

This paper takes the first important step to align OSM and knowledge graphs at the schema level using a novel neural method. 
In particular, we tackle tag-to-class alignment, i.e., we aim to identify OSM tags that convey class information and map them to the corresponding classes in the Wikidata knowledge graph and the DBpedia ontology. 
We present the Neural Class Alignment (\approach{}) model  - a novel instance-based neural approach that aligns OSM tags with the corresponding semantic classes in a knowledge graph.
\approach{} builds upon a novel shared latent space that aligns OSM tags and KG concepts and facilitates a seamless translation between them.
To the best of our knowledge, \approach{} is the first approach to align OSM and KGs at the schema level with a neural method.

Our contributions are as follows: 
\begin{itemize}
    \item We present \approach{} - a novel approach for class alignment for OSM and KGs. 
    \item We propose a novel shared latent space that fuses feature spaces from knowledge graphs and OSM in a joint model enabling simultaneous training of the schema alignment model on heterogeneous semantic and geographic sources.
    \item We develop an effective algorithm to extract tag-to-class alignments from the resulting model.
    \item The results of our evaluation demonstrate that the proposed \approach{} approach is highly effective and outperforms the baselines by up to 53 percentage points in terms of F1-score.
    \item As a result of the proposed \approach{} alignment method, we provide semantic annotations with Wikidata and DBpedia classes for over 10 million OSM entities. This result corresponds to a more than 400\% increase compared to currently existing annotations.
     \item We make our code and datasets publicly available and provide a manually annotated ground truth for the tag-to-class alignment of OSM tags with Wikidata and DBpedia classes\footnote{GitHub repository: \url{https://github.com/alishiba14/NCA-OSM-to-KGs}}.
\end{itemize}

%% file: example.tex
\begin{minipage}[t]{0.48\textwidth}
\centering
\footnotesize
Wikidata\\[0.5em]
$
\begin{array}{lll} 
\hline
\textbf{Subject} &  \textbf{Predicate} & \textbf{Object}\\
\hline
    Q3375 & label & Zugspitze \\
    
    Q3375 & coordinate &  47\degree25'N, 10\degree59'E \\
    Q3375 & parent peak & Q15127\\
    Q3375 & instance ~of & mountain

\end{array}
$
\end{minipage}
\hfill
\begin{minipage}[t]{0.5\textwidth}
\centering
\footnotesize
OpenStreetMap\\[0.5em]
$
\begin{array}{ll}
\hline
\textbf{Key} &  \textbf{Value}\\
\hline
    id & 27384190 \\
    \texttt{name}& Zugspitze \\
    \texttt{natural} & peak \\
    \texttt{summit:cross} & yes\\\\
\end{array}
$
\end{minipage}

%% file: 02_problem.tex
\section{Problem Statement}
\label{sec:problem}

In this section, we formalize the problem definition.
First, we formally define the concepts of an OSM corpus and a knowledge graph. 
An OSM corpus contains nodes representing geographic entities. Each node is annotated with an identifier, a location, and a set of key-value pairs known as tags.

\begin{definition}
\label{def:osm}
An \emph{OSM corpus} $\mathcal{C}  = (N, T)$ 
consists of a set of nodes $N$ representing geographic entities, and a set of tags $T$. 
Each tag $t \in T$ is represented as a key-value pair with the key $k \in K$ and a value $v \in V$: $t=\langle k,v \rangle$. 
A node $n \in N$,  $n=\langle i, l, T_{n} \rangle$ is represented as a tuple containing an identifier $i$, a geographic location $l$, and a set of tags $T_{n}\subset T$. 
\end{definition}

A KG contains real-world entities, classes, properties, and relations. 

\begin{definition}
A \emph{knowledge graph} $\mathcal{KG}= (E, C, P, L, F)$
consists of a set of entities $E$, a set of classes $C \subset E$, a set of properties $P$, a set of literals $L$,
and a set of triples $F \subseteq E \times P \times (E \cup L)$. 
\end{definition}
The entities in $E$ represent real-world entities and semantic classes.
The properties in $P$ represent relations connecting two entities, or an entity and a literal value.
An entity in a KG can belong to one or multiple classes.
An entity is typically linked to its class using the \texttt{rdf:type}, or an equivalent property.

\begin{definition}
A class of the entity $e\in E$ in the knowledge graph \\ $\mathcal{KG}= (E, C, P, L, F)$
is denoted as: $\textit{class(e)} = \{ c \in C ~| ~(e, \texttt{rdf:type}, c) \in F \}$.
\end{definition}

An OSM node and a KG entity referring to the same real-world geographic entity and connected via an identity link are denoted linked entities.

\begin{definition}
A \textit{linked entity} $(n, e) \in E_L$ is a pair of an OSM node $n=\langle i, l, T_n \rangle$, $n \in N$, and a knowledge graph entity $e \in E$ that correspond to the same real-world entity. 
In a knowledge graph, a linked entity is typically represented using a $(e, \texttt{owl:sameAs}, i)$ triple, where $i$ is the node identifier.
$E_L$ denotes the set of all linked entities in a knowledge graph.
\end{definition}

This paper tackles the alignment of tags that describe types of nodes in an OSM corpus to equivalent classes in a knowledge graph.
\begin{definition}
\textbf{Tag-to-class alignment:}
Given a knowledge graph $\mathcal{KG}$ and an OSM corpus $\mathcal{C}$, 
find a set of pairs  $ \textit{tag\_class}  \subseteq (T \times C)$ of OSM tags $T$ and the corresponding $\mathcal{KG}$ classes such that for each pair $(t,c) \in \textit{tag\_class} $  OSM nodes with the tag $t$ belong to the class $c$.
\end{definition}

%% file: 03_approach.tex
\section{Neural Class Alignment Approach}
\label{sec:approach}

An alignment of an OSM corpus with a knowledge graph can include several dimensions, such as entity linking, node classification (i.e., alignment of OSM nodes with the corresponding semantic classes in a knowledge graph), as well as alignment of schema elements such as keys/tags and the corresponding semantic classes. These dimensions can reinforce each other. For example, linking OSM nodes with knowledge graph entities and classification of OSM nodes into knowledge graph classes can lead to new schema-level alignments and vice versa. 
Our proposed \approach{} approach systematically exploits the existing identity links between OSM nodes and knowledge graph entities based on this intuition. \approach{} builds an auxiliary classification model and utilizes this model to align OSM tags with the corresponding classes in a knowledge graph ontology.

\begin{figure}[!t]
    \centering
    \includegraphics[width=0.75\textwidth]{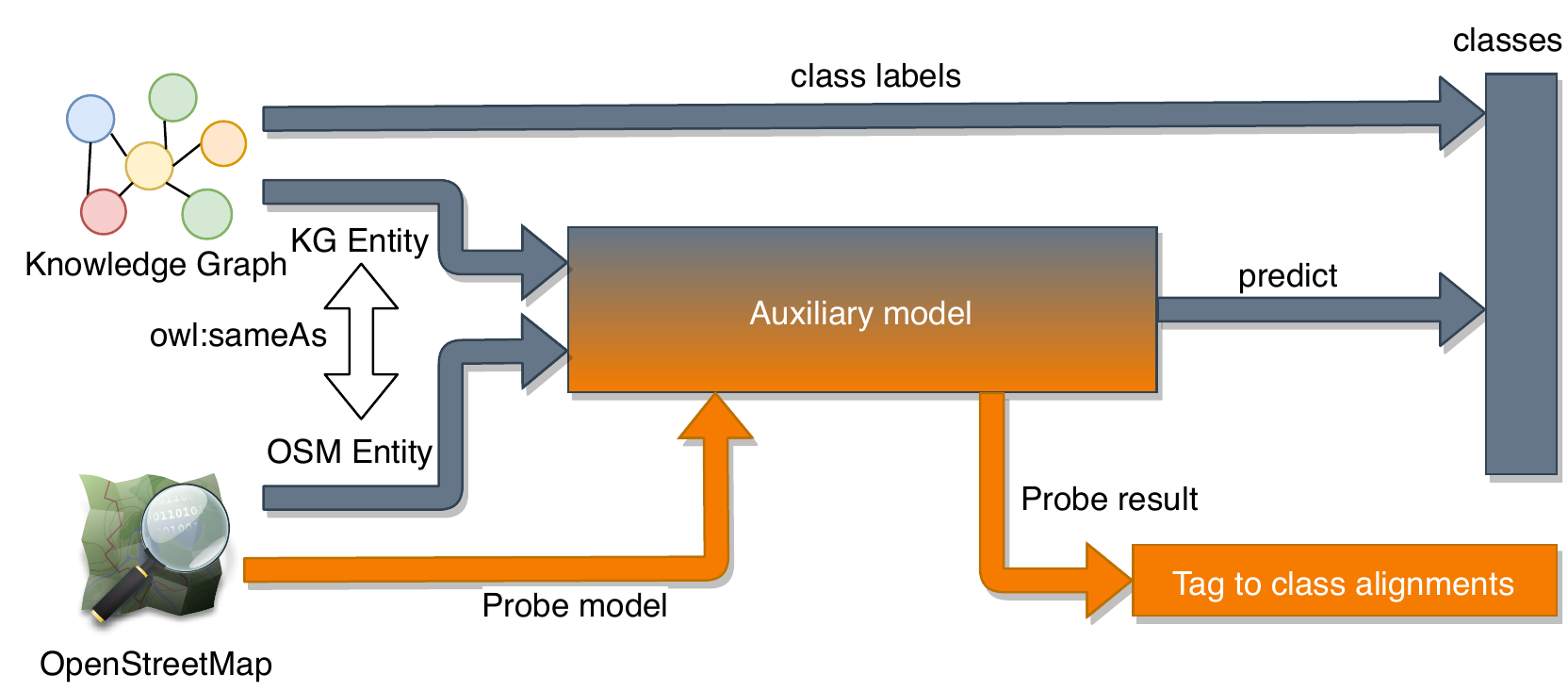}
    \caption{Overview of the \approach{} architecture. Grey color indicates the first step (training of the auxiliary classification model). Orange color indicates the second step, i.e., the extraction of tag-to-class alignments.}
    \label{fig:classification}
\end{figure}

\approach{} is an unsupervised two-step approach for tag-to-class alignment.
Fig. \ref{fig:classification} presents an overview of the proposed \approach{} approach architecture.
    First, we build an auxiliary neural classification model and train this model using linked entities in OSM and a KG. As a result, the model learns a novel shared latent space that \ed{aligns the feature spaces of OSM and a knowledge graph and} implicitly captures tag-to-class alignments. 
    Second, we systematically probe the resulting model to identify the captured alignments.

\subsection{Auxiliary Neural Classification Model}
\label{sec:class-model}

\begin{figure}[]
    \centering
    \includegraphics[width=0.75\textwidth]{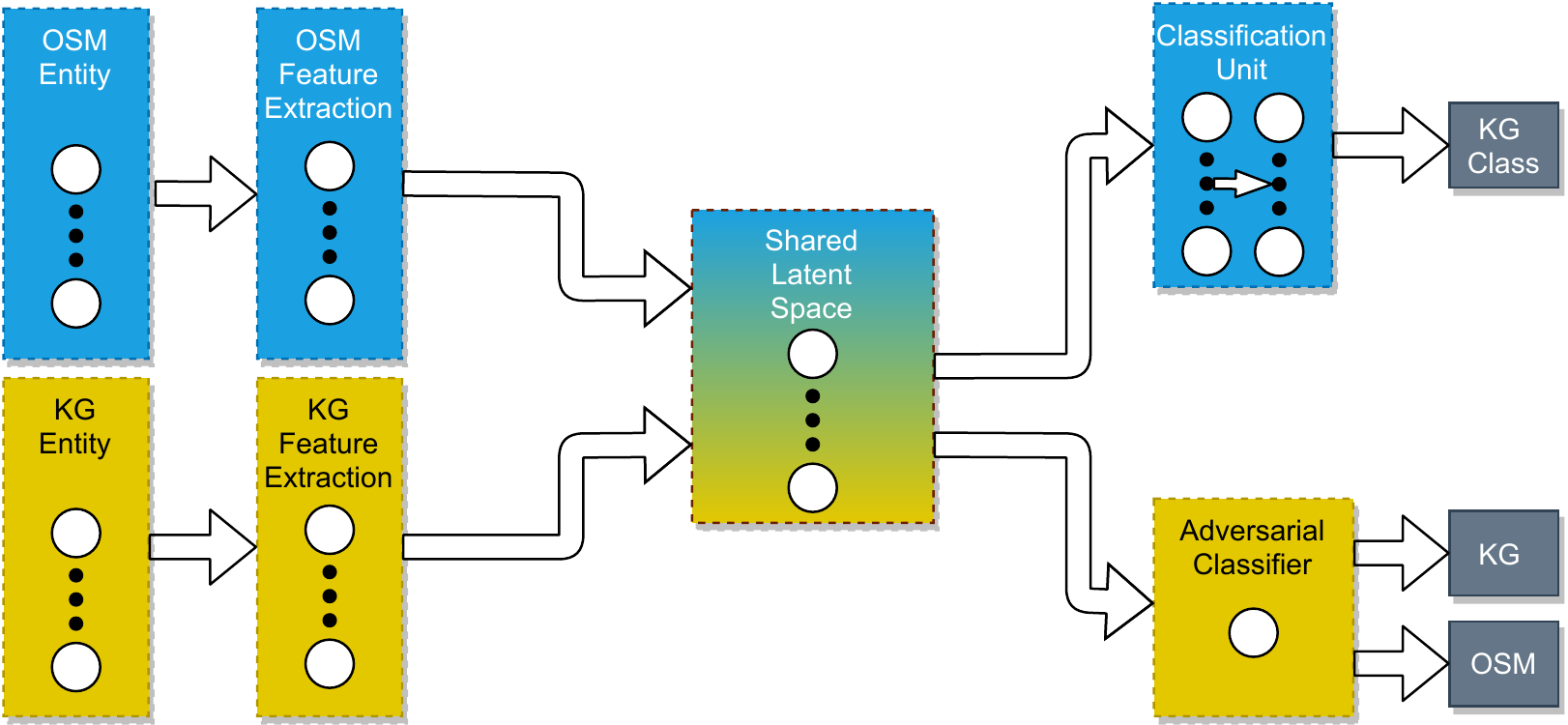}
    \caption{Neural architecture of the auxiliary classification model. Blue color indicates the KG classification component, yellow marks the adversarial entity discrimination component.}
    \label{fig:aux}
\end{figure}

In this step, we build a supervised auxiliary neural classification model for a dummy task of OSM node and KG entity classification. 
The model resulting from this step is later used for the tag-to-class alignment. 
Fig. \ref{fig:aux} presents the model architecture. 
The auxiliary classification model architecture consists of several components described below.

\textbf{OSM node representation.}
\ed{
We represent an OSM node as a binary vector in an $\mathbf{O}$-dimensional vector space. The space dimensions correspond to OSM tags or keys, and binary values represent whether the node includes the corresponding tag or key. 
The vector space dimensions serve as features for the classification model, such that we also refer to this space as the OSM feature space.
To select the most descriptive tags to be included as dimensions in the OSM feature space,
we filter out low-quality tags using OSM taginfo\footnote{OSM taginfo: \url{https://taginfo.openstreetmap.org/tags}}.
We include only the tags with an available description in the OSM wiki\footnote{OSM wiki: \url{https://wiki.openstreetmap.org/wiki/}} having at least 50 occurrences within OSM. 
For tags with infrequent values (e.g., literals), we include only the keys as dimensions.
As we aim to align geographic concepts and not specific entities, we do not include infrequent and node-specific values such as entity names or geographic coordinates in the representation.
For instance, the concept of ``mountain'' is the same across different geographic regions, 
such that the geographic location of entities is not informative for the schema alignment. 
}

\textbf{KG entity representation.}
\ed{
We represent a KG entity as a binary vector in a $\mathbf{V}$-dimensional vector space. The space dimensions correspond to the KG properties, 
and binary values represent whether the entity includes the corresponding property. 
The vector space dimensions serve as features for the classification model, such that we also refer to this space as the KG feature space.}
To select the most descriptive properties to be included in the KG feature space, we rank the properties based on their selectivity concerning the class and the frequency of property usage (\ed{i.e., the number of statements in the KG that assign this property to an entity}). 
Given a property $p$, we calculate its weight as:
    $weight(p,c) = n_{p,c} * log \frac{N}{c_{p}}$.

Here, $n_{p,c}$ denotes the number of statements in which the property $p$ is assigned to an entity of class $c$, 
$N$ denotes the total number of classes in a knowledge graph, and $c_p$ is the number of distinct classes that include the property $p$. 
For each class $c$, we select top-25 properties as features.
These properties are included as dimensions in the KG feature space.

\textbf{OSM \& KG feature extraction.}
The KG and OSM feature representations serve as input to the specific fully connected feature extraction layers: OSM feature extraction and KG feature extraction. 
The purpose of these layers is to refine the vector representations
obtained in the previous step.

\textbf{Shared latent space \& adversarial classifier.}
We introduce a novel \emph{shared latent space} that fuses the initially disjoint feature spaces of OSM and KG such that entities from both data sources are represented 
in a joint space similarly.
In addition to the training on OSM examples, shared latent space enables us to train our model on the KG examples. 
These examples provide the properties known to indicate class information \cite{DBLP:conf/semweb/PaulheimB13}.
The shared latent space component consists of a fully connected layer that receives the input from the OSM and KG feature extraction layers.
Following recent domain adaption techniques \cite{ganin2016domainadversarial}, we use an adversarial classification layer to align latent representations of KG and OSM entities.
The objective of the adversarial classifier is to discriminate whether the current training example is an OSM node or a KG entity, where the classification loss is measured as binary cross-entropy.

   $ BinaryCrossEntropy =-\frac {1}{n}\sum \limits _{i=1}^{n}{[ y_{i}\times \log {({\hat{y}_i})}} +{(1-y}_{i})\times \log{(1-{\hat{y}_i})} ]$,\\
\noindent 
where $n$ is the total number of examples, $y_{i}$ is the true class label, and $\hat{y}_i$ is the predicted class label.
Intuitively in a shared latent space, the classifier should not be able to distinguish whether a training example originates from OSM or a KG.
To fuse the initially disjoint feature spaces, we reverse the gradients from the adversarial classification loss: 
$\mathcal{L}_{adverse} = - BinaryCrossEntropy_{adverse}$.

\textbf{Classification Unit.}
To train the auxiliary classification model for the OSM nodes, we exploit linked entities. We label OSM nodes with semantic classes of equivalent KG entities. 
We use these class labels as supervision in the OSM node classification task. 
More formally, given a linked entity $(n,e) \in E_L$ the training objective of the model is to predict $\textit{class}(e)$ from $n$.
Analogously, the training objective for a KG entity $e$ is to predict the class label $\textit{class}(e)$ of this entity.

We utilize a 2-layer feed-forward network as a classification model.
In the last prediction layer of this network, each neuron corresponds to a class.
As an entity can be assigned to multiple classes, we use a sigmoid activation function and a binary cross-entropy loss 
to achieve multi-label classification:
   $ \mathcal{L}_{classifcation} =  BinaryCrossEntropy_{classification}$.
\noindent Finally, the joint loss function $\mathcal{L}$ of the network is given by 
   $\mathcal{L}= \mathcal{L}_{classifcation} + \mathcal{L}_{adverse}$.
In the training process, we alternate OSM and KG instances to avoid bias towards one data source. 

\subsection{Tag-to-Class Alignment}
\label{sec:tag-to-class-align}

In this step, we systematically probe the trained auxiliary classification model to extract the tag-to-class alignment.
The goal of this step is to obtain the corresponding KG class for a given OSM tag. 
Algorithm \ref{alg:cls-tag} details the extraction process.
First, we load the pre-trained auxiliary model $m$ (line 1) and initialize the result set (line 2).
We then probe the model with a given list of OSM tags $\mathcal{T}$ (line 3).
For a single tag $t \in \mathcal{T}$, we feed $t$ to the OSM input layer of the auxiliary model and compute the complete forward propagation of $t$ within $m$ (line 4).
We then extract the activation of the neurons of the last layer of the classification model before the sigmoid nonlinearity (line 5).
As the individual neurons in this layer directly correspond to KG classes
we expect that the activation of the specific neurons quantifies the likeliness that the tag $t$ corresponds to the respective class.
For each activation of a specific neuron $a$ that is above the alignment threshold $th_a$ (line 6-7), we extract the corresponding class $c$ and add this class to the set of alignments (line 8).
We determine the threshold value experimentally, as described later in Section \ref{sec:ev-conf-threshold}. 
As an OSM tag can have multiple corresponding classes, we opt for all matches above the threshold value.
Finally, the resulting set $align$ constitutes the inferred tag-to-class alignments.

\begin{algorithm}[!t]
\footnotesize
\begin{flushleft}
\begin{tabular}{lll}
\footnotesize

     \texttt{Input:} & $m$ & Trained auxiliary model  \\
                     & $\mathcal{T}$ & List of OSM tags \\
                     & $th_a$ & Alignment threshold \\
     \texttt{Output:} & $align \subseteq (T \times C)$ & Extracted alignment of tags and classes
\end{tabular}
\end{flushleft}
\hrule
\begin{algorithmic}[1]
\footnotesize
    \STATE \texttt{load}($m$)
    \STATE $align \Leftarrow \emptyset$
    \FORALL{$t \in \mathcal{T}$}
        \STATE \texttt{forward\_propagation(t, m)}
        \STATE $activations \Leftarrow \texttt{extract\_activations}(m)$
        \FORALL{ $a \in activations$}
            \IF {$a > th_a$}
                \STATE $align \Leftarrow align \cup \{(t, \texttt{class}(a))\}$
            \ENDIF            
        \ENDFOR
    \ENDFOR
    \RETURN $align$
\end{algorithmic}
\caption{Extract Tag-to-Class Alignment}
\label{alg:cls-tag}
\end{algorithm}

\subsection{Illustrative Example}

We illustrate the proposed \approach{} approach at the example of the ``Zugspitze" mountain introduced in Section \ref{sec:introduction}. 
We create the representation of the Wikidata object ``Q3375" in the KG feature space 
by creating a binary vector that has ones in the dimensions that correspond to the properties this entity contains such as, \texttt{label, coordinate, parentpeak} and zeros otherwise. 
Note that \texttt{instance of} predicate is not included in the feature space as it represents the class label.
Similarly, we encode the OSM node with the id ``27384190" in the OSM feature space by creating a vector that includes 
\texttt{name, natural=peak, summit:cross} as ones and has zeros in all other dimensions. 
As described above, we use frequent key-value pairs such as \texttt{natural=peak} as features, whereas for the infrequent key-value pairs, such as \texttt{name=Zugspitze}, we use only the key (i.e., \texttt{name}) as a feature. 
The KG and OSM features spaces are then aligned in the shared latent space.
To form this space, we train the auxiliary classification model that learns to outputs the correct class labels, such as ``mountain".
In the last prediction layer of this model, each neuron corresponds to a class.
After the training is completed, we can prob the classification model with a single tag such as \texttt{natural=peak}. 
The activation of the neurons in the prediction layer correspond to the predicted tag-to-class mapping.
We output all classes with the activation values above the threshold $th_a$ (here: ``mountain").

%% file: 04_setup.tex
\section{Evaluation Setup}
\label{sec:ev-setup}
This section introduces the evaluation setup in terms of datasets, ground truth generation, baselines, and evaluation metrics. All experiments were conducted on an AMD Opteron 8439 SE processor @ 2.7 GHz and 252 GB of memory.

\subsection{Datasets}
\label{sec:ev-datasets}

We carry out our experiments on OSM, Wikidata \cite{articlewikidata}, and DBpedia \cite{articledbpedia} datasets.

\textbf{Knowledge graphs:} 
A sufficient number of linked entities and distinct classes is essential to train the proposed neural model and to achieve a meaningful schema alignment. 
We systematically rank European countries according to the number of linked entities between OSM and knowledge graphs and choose the top-4 countries having at least ten distinct classes in the linked entity set.
Based on these criteria, we select the Wikidata datasets for France, Germany, Great Britain, and Russia and the DBpedia datasets for France, Germany, Great Britain, and Spain. 
Although over 100,000 entity links between Russian DBpedia and OSM exist, most of these entities belong to only two classes. 
Hence, we omit Russian DBpedia from our analysis.
In our experiments, we consider Wikidata and DBpedia snapshots from March 2021. 
We only consider geographic entities, i.e., 
the entities with valid geographic coordinates.

\textbf{OpenStreetMap:} 
We extract OSM data for France, Germany, Great Britain, Spain, and Russia.
To facilitate evaluation, we only consider OSM nodes which include links to knowledge graphs. 
The number of entities assigned to specific knowledge graph classes follows a power-law distribution. 
We select the classes with more than 100 entities (i.e., 3\% of classes in Wikidata) to facilitate model training. 
Note that some KG entities are linked to more than one OSM node, such that the number of OSM nodes and KG entities in the dataset differ.

\subsection{Ground Truth Creation}
\label{sec:ev:groundTruth}

For Wikidata, we start the creation of our ground truth based on the ``OpenStreetMap tag or key" Wikidata property\footnote{Wikidata ``OpenStreetMap tag or key" property: \url{https://www.wikidata.org/wiki/Property:P1282}}. This property provides a link between a Wikidata class and the corresponding OSM tag.
However, this dataset is incomplete and lacks some language-specific classes as well as superclass and subclass relationships based on our manual analysis.  
We manually extended the ground truth by checking all possible matches obtained by the proposed \approach{} approach and all baseline models used in the evaluation and added all correct matches to our ground truth. 
For DBpedia, we constructed the ground truth manually by labeling all combinations $(T \times C)$ of OSM tags $t$ and $\mathcal{KG}$ classes $C$ in our dataset. For both KGs, we take region-specific matches (``Ortsteil" vs. ``District") as well as subclass and superclass relations (e.g., ``human settlement"  vs. ``city/village") into consideration.

\subsection{Baselines}
\label{sec:ev-baselines}
As OSM has a flat schema, ontology alignment methods that utilize structural features are not applicable. Therefore, we utilize schema alignment methods based on the schema element names and methods developed for tabular data as baselines. 
To fit our data to the baselines, we convert our OSM (source) data and KG (target) data into a tabular format. For OSM, we use the tags and keys as columns and convert each node into a row. Similarly, for KGs, the properties and classes are converted into columns, and the entities form the rows.
Overall, we evaluate our proposed method 
against the following baselines:

\textbf{Cupid}: 
Cupid \cite{madhavan2001generic} matches schema elements based on element names, structure, and data types. Cupid is a 2-phase approach. The first phase calculates the lexicographic similarity of names and data types. The second phase matches elements using the structural similarity based on the element proximity in the ontology and the hierarchical position. As the OSM schema is flat, we consider a flat hierarchy, where the OSM table is the root and all columns are child nodes. The final Cupid score is the average similarity between the two phases. 

\textbf{Levenshtein Distance (LD)}: 
The Levenshtein distance (edit distance) is a string-based similarity measure used to match ontology elements lexicographically. 
The Levenshtein distance between two element names is calculated as the minimal number of edits needed to transform one element name to obtain the other. The modifications include addition, deletion, or replacement of characters \cite{inproceedingsunal}. 
We calculate the Levenshtein distance between all pairs of class names and tags and accept all pairs with a distance lower than a matching threshold $th_l \in [0, 1]$. 
We apply an exhaustive grid search to optimize the value of $th_l$ for each dataset and report the highest resulting F1-scores.

\textbf{EmbDi}: 
EmbDi \cite{10.1145/3318464.3389742} is an algorithm for schema alignment and entity resolution.
The algorithm maps table rows to a directed graph based on rows, columns, and cell values. EmbDi infers column embeddings by performing random walks on the graph. The random walks form sentences that constitute an input to a Word2Vec model. Finally, the similarity of the two columns is measured as the cosine similarity of the respective embeddings.

\textbf{Similarity Flooding (SF)}: 
Similarity Flooding \cite{994702} transforms a data table into a directed labeled graph in which the nodes represent table columns. The weights of graph edges represent the node similarity, initialized using string similarity of the column names. The algorithm refines the weights by iteratively propagating similarity values along the edges. Each pair of nodes connected with a similarity value above the matching threshold forms an alignment. The algorithm filters out specific matches such as data types and schema elements. We experimentally select a threshold value of 0 to increase recall, as at higher threshold values, the algorithm obtains no matches.

\subsection{Metrics}
\label{sec:ev-metrics}

The standard evaluation metrics for schema alignment are precision, recall, and F1-score computed against a reference alignment (i.e., ground truth).
We evaluate the $kg_{class}$ mappings as pairs, where each pair consists of one tag and one class (tag-to-class alignment).
\textbf{Precision} is the fraction of correctly identified pairs among all identified pairs.
\textbf{Recall} is the fraction of correctly identified pairs among all pairs in the reference alignment.
\textbf{F1-score} is the harmonic mean of recall and precision. We consider the F1-score to be the most relevant metric since it reflects both recall and precision.

%% file: 05_evaluation.tex
\section{Evaluation}
\label{sec:ev-results}

The evaluation aims to assess the performance of the proposed \approach{} approach for tag-to-class alignment in terms of precision, recall, and F1-score. Furthermore, we aim to analyze the influence of the confidence threshold and the impact of the shared latent space on the alignment performance.
Note that we do not evaluate the artificial auxiliary classification task. Instead, we evaluate the utility of the auxiliary model in the overall schema alignment task.

\subsection{Tag-to-Class Alignment Performance}
\label{sec:eval-tag-to-class}
Table \ref{tab:class_to_tag} and Table \ref{tab:class_to_tag_wiki} summarize the performance results of the baselines as well as our proposed \approach{} approach with respect to precision, recall and F1-score for tag-to-class alignment of OSM tags to Wikidata and DBpedia classes, respectively. 

The proposed \approach{} approach outperforms the baselines in terms of precision and F1-score on all datasets. 
On Wikidata, we achieve up to 30 percent points F1-score improvement and 19.75 percent points on average compared to the best baseline. On DBpedia, we achieve up to 53 percent points F1-score improvement and 37.5 percent points on average.
As OSM lacks a hierarchical structure, limiting structural comparison, 
most of the applicable baselines build on the name comparison. 
Here, the heterogeneity of OSM tags limits the precision of the baselines substantially. 
Compared to the other baselines, Levenshtein distance and Cupid obtain the highest F1-scores for Wikidata and  DBpedia, respectively,  even though the absolute values achieved by these baselines are relatively low. 
SF and EmbDI obtain only low similarity values, resulting in low precision. 
An increase of the confidence threshold for these baselines leads to zero matches.

\begin{table}[t]
    \caption{Tag-to-class alignment performance 
    for OSM tags to Wikidata classes.}
\input{tables/class_to_tag}

    \label{tab:class_to_tag}
\end{table}

\begin{table}[t]
    \caption{Tag-to-class alignment performance 
    for OSM tags to DBpedia classes.}
\input{tables/class_to_tag_wikipedia}

    \label{tab:class_to_tag_wiki}
\end{table}

\begin{table}[t]
\centering
\caption{Tag-to-class alignments obtained using \approach{} approach.}
\resizebox{\textwidth}{!}{
\begin{tabular}{ccccc}

\textbf{}                                                                                      & France                                                        & Germany                                                        & Great Britain                                                          & Russia                                                     \\ \hline

\begin{tabular}[c]{@{}c@{}} Wikidata\end{tabular}               & \begin{tabular}[c]{@{}c@{}}amenity=bicycle\_rental:\\ bicycle-sharing station \end{tabular} & \begin{tabular}[c]{@{}c@{}}amenity=cinema:\\ movie theater\end{tabular} & \begin{tabular}[c]{@{}c@{}}railway=station:\\ railway station\end{tabular} & \begin{tabular}[c]{@{}c@{}}station=subway:\\ metro station\end{tabular} \\ \hline

\textbf{}                                                                                      & France                                                        & Germany                                                        & Great Britain                                                          & Spain                                                     \\ \hline

\begin{tabular}[c]{@{}c@{}}DBpedia\end{tabular}   & \begin{tabular}[c]{@{}c@{}}railway=station:   \\ Place \end{tabular} & \begin{tabular}[c]{@{}c@{}}place=municipality:  \\ Place
\end{tabular} & \begin{tabular}[c]{@{}c@{}}place=hamlet:  \\ Place\end{tabular} & \begin{tabular}[c]{@{}c@{}}railway=station:   \\ ArchitecturalStructure\end{tabular} \\ \hline
\end{tabular}
}
\label{tab:tag-to-class-stat-Wikidata}
\end{table}

\ed{
We observe performance variations across countries and knowledge graphs, with French Wikidata and DBpedia achieving the highest F1-scores in comparison to other countries. These variations can be explained by the differences in the dataset characteristics, such as the number of links, the number of entities per class, and the number of unique tags and classes per country. These characteristics vary significantly across the datasets.
Furthermore, the number of classes per entity varies. On average, Wikidata indicates one class per entity (i.e., the most specific class), whereas DBpedia indicates three classes per entity on average (i.e., the specialized as well as more generic classes at the higher levels of the DBpedia ontology). This property makes the model trained on the DBpedia knowledge graph more confident regarding the generic classes compared to the specialized classes, such that generic classes obtain higher F1-scores.
Our observations indicate that it is desirable to obtain more training examples that align entities with more specific classes, such as in the Wikidata dataset.
}

Table \ref{tab:tag-to-class-stat-Wikidata}  illustrates the most confident tag-to-class alignments in terms of the obtained model activations using the \approach{} approach. As discussed above, Wikidata alignments with high confidence scores are more specific than those obtained with the DBpedia datasets.

\subsection{Influence of the Shared Latent Space}
\label{sec:ev-adversarial}

Table \ref{tab:cls-to-tag-adverse-short}
summarizes the performance of the proposed \approach{} approach and \approach{} without the shared latent space for tag-to-class alignment of OSM with Wikidata and DBpedia, respectively.

We observe that the shared latent space helps 
to achieve an increase in F1-score of 22 percentage points and 12 percentage points for Wikidata and DBpedia, respectively.

\begin{table}[!t]
    \caption{Tag-to-class alignment performance for Wikidata and DBpedia.}
\input{tables/table-shared-space-avg}

    \label{tab:cls-to-tag-adverse-short}
\end{table}

Compared to the Wikidata datasets, we observe smaller improvements on DBpedia datasets.
DBpedia has an imbalance between the tags and classes that results in many-to-one alignments between tags and classes, 
where one class corresponds to several different tags.
For example, in all DBpedia datasets, the \emph{place} and \emph{populatedPlace} are frequently occurring classes for various tags such as \emph{tourism=museum, place=village, place=town}.
In such a case, DBpedia properties add less focused information to the matching process. 
Furthermore, we observe high F1-score of the proposed \approach{} model without the shared latent space on DBpedia dataset.
Intuitively, further improving these high scores is more difficult than improving the comparable low scores observed on Wikidata (e.g., 0.4  F1-score on Wikidata).
In summary, the shared latent space improves the matching performance. We observe the highest improvements on Wikidata.

\subsection{Confidence Threshold Tuning}
\label{sec:ev-conf-threshold}
\begin{figure}[!t]
    \centering
    \begin{minipage}[t]{\textwidth}
        \centering
        \includegraphics[width=0.45\textwidth]{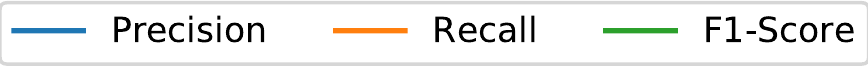}
    \end{minipage}\vspace{3pt}
    \begin{minipage}[t]{\textwidth}
    \begin{subfigure}{0.2588\textwidth}
        \includegraphics[width=\textwidth]{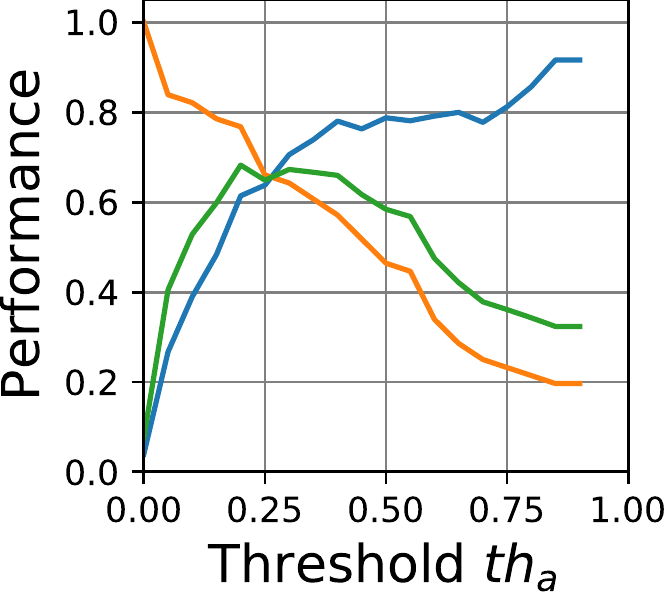}
        \caption{France}
    \end{subfigure}
    \begin{subfigure}{0.2337\textwidth}
        \includegraphics[width=\textwidth]{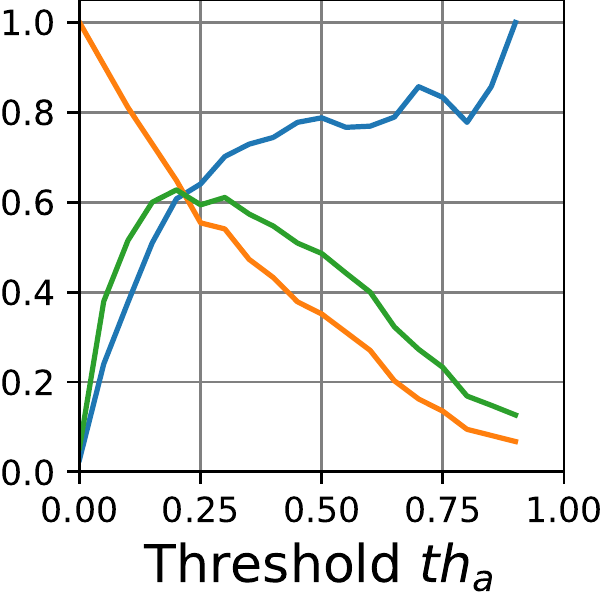}
        \caption{Germany}
    \end{subfigure}
    \begin{subfigure}{0.2337\textwidth}
        \includegraphics[width=\textwidth]{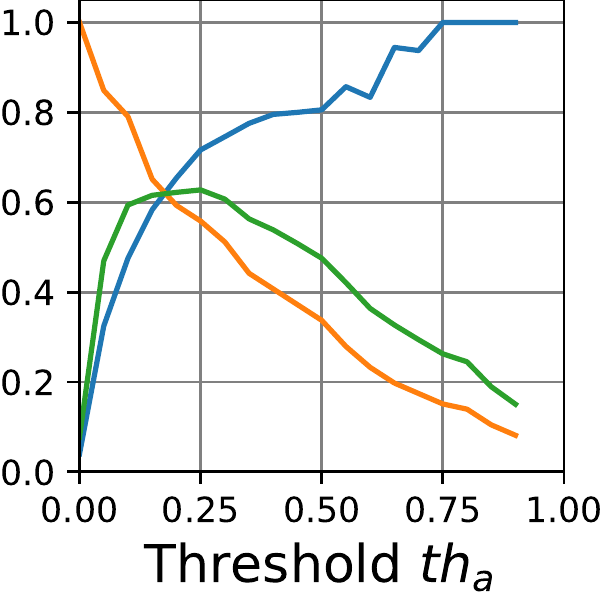}
        \caption{Great Britain}
    \end{subfigure}
    \begin{subfigure}{0.2337\textwidth}
        \includegraphics[width=\textwidth]{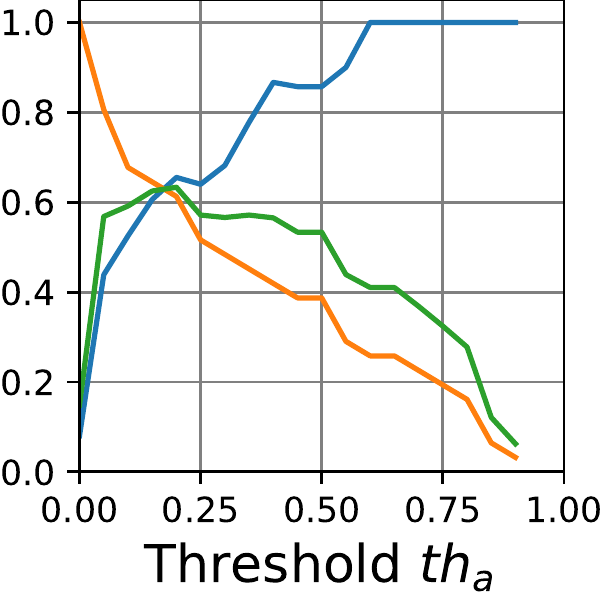}
        \caption{Russia}
    \end{subfigure}
    \end{minipage}

    \caption{Precision, recall, and F1-score vs. the confidence threshold for Wikidata.}
    \label{conf-cls-to-tag-wikidata}
\end{figure}

\begin{figure}[!t]
     \centering
         \begin{minipage}[t]{\textwidth}
         \centering
         \includegraphics[width=0.45\textwidth]{Images/legend_Horizontal_crop.pdf}
         \end{minipage}\vspace{3pt}
     \begin{minipage}[t]{\textwidth}
     \begin{subfigure}{0.2588\textwidth}
         \includegraphics[width=\textwidth]{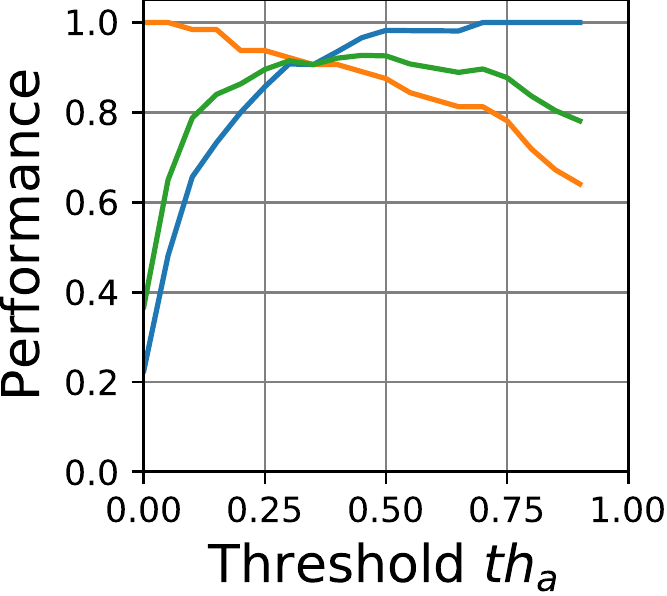}
         \caption{France}
     \end{subfigure}
     \begin{subfigure}{0.2337\textwidth}
         \includegraphics[width=\textwidth]{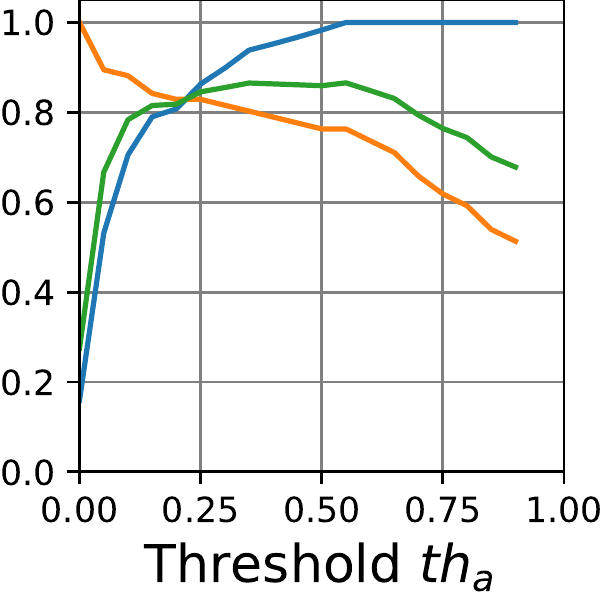}
         \caption{Germany}
     \end{subfigure}
     \begin{subfigure}{0.2337\textwidth}
         \includegraphics[width=\textwidth]{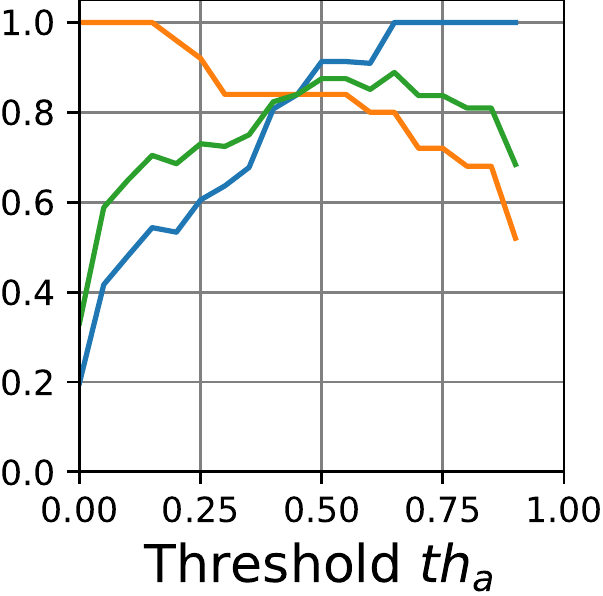}
         \caption{Great Britain}
     \end{subfigure}
     \begin{subfigure}{0.2337\textwidth}
         \includegraphics[width=\textwidth]{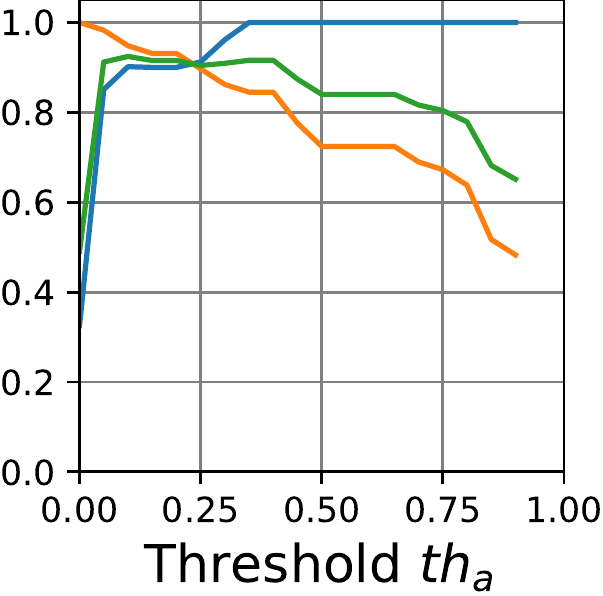}
         \caption{Spain}
     \end{subfigure}
     \end{minipage}

     \caption{Precision, recall, and F1-score vs. the confidence threshold for DBpedia.}
     \label{conf-cls-to-tag-DBpedia}
\end{figure}

We evaluate the influence of the confidence threshold value $th_a$ on the precision, recall, and F1-score.
The threshold $th_a$ indicates the minimum similarity at which we align a tag to a class.
Fig. \ref{conf-cls-to-tag-wikidata} and \ref{conf-cls-to-tag-DBpedia} 
present the alignment performance with respect to $th_a$ for Wikidata and DBpedia.
As expected, we observe a general trade-off between precision and recall, whereas higher values of $th_a$ result in higher precision and lower recall. 
In our experiments, we select the confidence threshold of $th_a=0.25$ and $th_a=0.4$ for Wikidata and DBpedia, respectively, as these values allow us to balance precision and recall.
The threshold can be further tuned for specific geographic regions.

\subsection{Alignment Impact}
\label{sec:ev-impact}

To assess the impact of \approach{}, we compare the number of OSM entities that can be annotated with semantic classes using the alignment discovery by \approach{} with the number of entities that are linked to a KG in the currently existing datasets. 
For Wikidata, we observe 2,004,510 linked OSM entities and 10,163,762 entities that can be annotated with semantic classes using \approach{}. This result corresponds to an increase of 407.04\% of entities with semantic class information.
For DBpedia, we observe 1,396,378 linked OSM entities and 8,301,450 entities that can be annotated with semantic classes. This result corresponds to an increase of 494.5\% of entities with semantic class information.
We make the resulting dataset available to facilitate reproducibility and further research.

%% file: tables/class_to_tag.tex
\resizebox{\textwidth}{!}{
\begin{tabular}{l@{\quad}rrc@{\hskip 1em}rrc@{\hskip 1em}rrc@{\hskip 1em}rrc@{\hskip 1em}rrc}
\multirow{2}{*}{\raisebox{-\heavyrulewidth}{\textbf{Approach}}} &               \multicolumn{3}{c}{\texttt{France}} 
		& \multicolumn{3}{c}{\texttt{Germany}} 
		& \multicolumn{3}{c}{\texttt{Great Britain}} 
        	& \multicolumn{3}{c}{\texttt{Russia}} 
        	& \multicolumn{3}{c}{\texttt{Average}}\\
\cmidrule(l{0pt}r{6pt}){2-4} \cmidrule(l{0pt}r{6pt}){5-7} \cmidrule(l{0pt}r{6pt}){8-10} \cmidrule(l{0pt}r{0pt}){11-13} \cmidrule(l{0pt}r{0pt}){14-16} 
    & Precision & Recall & F1 
    &  Precision & Recall & F1  
       &  Precision & Recall & F1  
    &  Precision & Recall & F1
    &  Precision & Recall & F1\\  
\midrule

\textsc{Cupid}   	&0.06 & 1 & 0.12
               			&0.03 & 0.70 & 0.06
               			&0.07 & 1 & 0.14 	
                		&0.08 & 0.80 & 0.15
                		& 0.08&0.88&0.12\\

\textsc{LD}   	&0.45 & 0.28 & 0.35
               			&0.65 & 0.34 & 0.44
               			&0.54 & 0.37 & 0.44
                		&0.64 & 0.34 & 0.45 
                		& 0.58 & 0.33 & 0.42\\

\textsc{EmbDi}   	&0.03 & 1 & 0.06
               			&0.02 & 1 & 0.03
               			&0.04 & 1 & 0.06
                		&0.02 & 1 & 0.03 
                		& 0.03 & 1 & 0.05\\
                		
\textsc{SF}   	&0.03 & 1 & 0.06
               			&0.02 & 1 & 0.03
               			&0.01 & 1 & 0.03
                		&0.02 & 1 & 0.03
                		& 0.02 & 1 & 0.04\\
                
\midrule

\textsc{\approach}	&0.63 & 0.66 & 0.65 
                    &0.59 & 0.65 & 0.61
                    &0.71 & 0.56 & 0.63 
                    &0.64 &	0.51 & 0.58
                    & 0.64 & 0.60 & 0.62\\

     \bottomrule
\end{tabular}}

%% file: tables/class_to_tag_wikipedia.tex
\resizebox{\textwidth}{!}{
\begin{tabular}{l@{\quad}rrc@{\hskip 1em}rrc@{\hskip 1em}rrc@{\hskip 1em}rrc@{\hskip 1em}rrc}
\multirow{2}{*}{\raisebox{-\heavyrulewidth}{\textbf{Approach}}} &               \multicolumn{3}{c}{\texttt{France}} 
		& \multicolumn{3}{c}{\texttt{Germany}} 
		& \multicolumn{3}{c}{\texttt{Great Britain}} 
		& \multicolumn{3}{c}{\texttt{Spain}}
		& \multicolumn{3}{c}{\texttt{Average}}\\
\cmidrule(l{0pt}r{6pt}){2-4} \cmidrule(l{0pt}r{6pt}){5-7} \cmidrule(l{0pt}r{6pt}){8-10}  \cmidrule(l{0pt}r{6pt}){11-13} \cmidrule(l{0pt}r{6pt}){14-16}
    & Precision & Recall & F1 
    &  Precision & Recall & F1  
       &  Precision & Recall & F1  
       &  Precision & Recall & F1  
       & Precision & Recall & F1
       \\  
\midrule

\textsc{Cupid}   	&0.32 & 1 & 0.48
               			&0.18 & 1 & 0.31
               			&0.41 & 1 & 0.58
               			&0.44 & 1 & 0.63
               			&0.33&1&0.50\\

\textsc{LD}   	&0.31 & 0.57 & 0.41
               			&0.32 & 0.37 & 0.34
               			&0.73 & 0.46 & 0.57
               			&0.34 & 0.94 & 0.50
               			&0.42&0.58&0.45\\

\textsc{EmbDi}   	&0.16 & 1 & 0.28
               			&0.09 & 1 & 0.17
               			&0.29 & 1 & 0.45
               			&0.24 & 1 & 0.38
               			&0.20&1&0.32\\
               
\textsc{SF}          	&0.14 & 1 & 0.27
               			&0.10 & 1 & 0.18
               			&0.27 & 1 & 0.42
               			&0.24 & 1 & 0.39
               			&0.19&1&0.31\\
                
\midrule

\textsc{\approach}	&0.95 & 0.90 & 0.92 
                    &0.96 & 0.79 & 0.87
                    &0.81 & 0.84 & 0.83 
                    & 1 &	0.84 & 0.91
                    &0.93&0.84&0.88\\

     \bottomrule
\end{tabular}}

%% file: tables/table-shared-space-avg.tex
\centering{
\resizebox{0.85\textwidth}{!}{
\centering
\begin{tabular}
{l@{\quad}rrc@{\hskip 1em}rrc@{\hskip 1em} rrc@{\hskip 1em}}
\multirow{2}{*}
{\raisebox{-\heavyrulewidth}
Approach} 
		& \multicolumn{3}{c}{Avg. Wikidata} 
		& \multicolumn{3}{c}{Avg. DBpedia}\\
\cmidrule(l{0pt}r{6pt}){2-4} \cmidrule(l{0pt}r{6pt}){5-7} \cmidrule(l{0pt}r{6pt}){8-10} \ 
       &  Precision & Recall & F1
       &  Precision & Recall & F1\\  
\midrule
\textsc{\approach} w/o shared latent space   	
               			&0.60&0.33&0.40 
                		&0.69 & 0.87 & 0.76 
                		\\ 

\midrule

\textsc{\approach}	
                    &0.64&0.60&0.62 
                    &0.93 &	0.84 & 0.88 
                    \\ 

     \bottomrule
\end{tabular}
}
}

%% file: 06_related_work.tex
\section{Related Work}
\label{sec:related}

This work is related to 
ontology alignment, alignment of tabular data, feature space alignment, and link discovery.

\textbf{Ontology Alignment.}
\label{sec:rel-ontology-alignment}
Ontology alignment (also ontology matching) aims to establish correspondences between the elements of different ontologies. 
The efforts to interlink open semantic datasets and benchmark ontology alignment approaches have been driven by the W3C SWEO Linking Open Data community project\footnote{\url{https://www.w3.org/wiki/SweoIG/TaskForces/CommunityProjects/LinkingOpenData}}
and the Ontology Alignment Evaluation Initiative (OAEI)\footnote{OAEI evaluation campaigns: \url{http://oaei.ontologymatching.org}} \cite{AlgergawyFFFHHJ19}.
Ontology alignment is conducted at the element-level, and structure-level \cite{Otero-CerdeiraRG15}.
The element-level alignment typically uses natural language descriptions of the ontology elements, such as labels and definitions. Element-level alignment adopts string similarity metrics such as, e.g., edit distance.
Structure-level alignment exploits the similarity of the neighboring ontology elements, including the taxonomy structure, as well as shared instances \cite{NgoBT13}.
Element-level and structure-level alignment have also been adopted to align ontologies with relational data \cite{DemidovaON13} and tabular data \cite{ZhangB20}.
Jiménez-Ruiz et al. \cite{abs-2003-05370} divided the alignment task into independent, smaller sub-tasks, aiming to scale up to very large ontologies. 
Machine learning has been widely adopted for ontology alignment. In the GLUE architecture \cite{Doan2004}, semantic mappings are learned in a semi-automatic way while \cite{Nkisi-OrjiWMHH18}  proposed a matching system that integrates string-based and semantic similarity features.
More recently, more complex approaches using deep neural networks have been used for ontology alignment and schema matching.
The proposed architectures include convolutional neural networks \cite{Bento2020OntologyMU}
and stacked autoencoders \cite{xiang-etal-2015-ersom}.
The lack of a well-defined ontology of OSM hinders the application of ontology alignment approaches to OSM data.
In contrast, our instance-based \approach{} approach presented in this paper enables an effective alignment of tags to classes.

\textbf{Tabular Data Alignment.}
Another branch of research investigated the schema alignment of tabular data \cite{DBLP:journals/vldb/RahmB01}.
Cappuzzo et al. \cite{10.1145/3318464.3389742} proposed EmbDi approach, which generates the graph structure for the tabular data, builds sentences from the graph, and generates embeddings to find similarity between the schema elements. 
Cupid \cite{madhavan2001generic} matches schema elements based on element names, structure, and data types.
Similarity Flooding \cite{994702} transforms a table into a directed labeled graph in which nodes represent columns to compute similarity values iteratively.
We employ the EmbDi, Cupid, and Similarity Flooding algorithms as baselines for our evaluation.

\textbf{Feature Space Alignment.}
\label{sec:rel-entity-classification}
Recently, various studies investigated the alignment of feature spaces extracted from different data sources.
Application domains include computer vision \cite{6751479} and machine translation \cite{lample2018word}.
Ganin et al. \cite{ganin2016domainadversarial} proposed a neural domain adaptation algorithm that considers labeled data from a source domain and unlabeled data from a target domain. 
While this approach was originally used to align similar but different distributions of feature spaces,
we adopt the gradient reversal layer proposed in \cite{ganin2016domainadversarial} to fuse information from the disjoint features spaces of OSM and KGs, not attempted previously.

\textbf{Link Discovery.} Link Discovery is the task of identifying semantically equivalent resources in different data sources \cite{DBLP:journals/semweb/NentwigHNR17}. 
Nentwig et al. \cite{DBLP:journals/semweb/NentwigHNR17} provide a recent survey of link discovery frameworks with prominent examples, including Silk \cite{www2009227} and LIMES \cite{Ngomo:2011:LTA:2283696.2283783}. 
In particular, the Wombat algorithm, integrated within the LIMES framework \cite{10.1007/978-3-319-58068-5_7}, is a state-of-the-art approach for link discovery in knowledge graphs. 
Specialized approaches \cite{TEMPELMEIER2021349} focus on the discovery links between OSM and knowledge graphs.
We build on existing links between OSM and knowledge graphs to align knowledge graph classes to OSM tags in this work.

%% file: 07_conclusion.tex
\section{Conclusion}
\label{sec:conclusion}

In this paper, we presented \approach{} - the first neural approach for tag-to-class alignment between OpenStreetMap and knowledge graphs. We proposed a novel shared latent space that seamlessly fuses features from knowledge graphs and OSM in a joint model and makes them simultaneously accessible for the schema alignment.
Our model builds this space as the core part of a neural architecture incorporating an auxiliary classification model and an adversarial component. Furthermore, we proposed an effective algorithm that extracts tag-to-class alignments from the resulting shared latent space with high precision. Our evaluation results demonstrate that \approach{} is highly effective and outperforms the baselines by up to 53 percentage points in F1-score. We make our code and manually annotated ground truth data publicly available to facilitate further research.

%% file: main.bbl
\begin{thebibliography}{10}
\providecommand{\url}[1]{\texttt{#1}}
\providecommand{\urlprefix}{URL }
\providecommand{\doi}[1]{https://doi.org/#1}

\bibitem{AlgergawyFFFHHJ19}
Algergawy, A., et~al.: Results of the ontology alignment evaluation initiative
  2019. In: Proc. of the OM-2019. {CEUR} Workshop Proceedings, vol.~2536 (2019)

\bibitem{articledbpedia}
Auer, S., Bizer, C., Kobilarov, G., Lehmann, J., Cyganiak, R., Ives, Z.:
  {DBpedia: A Nucleus for a Web of Open Data}. In: ISWC 2007. Springer (2007)

\bibitem{Bento2020OntologyMU}
Bento, A., Zouaq, A., Gagnon, M.: Ontology matching using convolutional neural
  networks. In: LREC (2020)

\bibitem{10.1145/3318464.3389742}
Cappuzzo, R., Papotti, P., Thirumuruganathan, S.: Creating embeddings of
  heterogeneous relational datasets for data integration tasks. In: SIGMOD'20
  (2020)

\bibitem{DemidovaON13}
Demidova, E., Oelze, I., Nejdl, W.: Aligning freebase with the {YAGO} ontology.
  In: Proceedings of CIKM 2013. pp. 579--588. {ACM} (2013)

\bibitem{Doan2004}
Doan, A., Madhavan, J., Domingos, P., Halevy, A.: Ontology matching: A machine
  learning approach. Handbook on Ontologies in Information Systems  (07 2003)

\bibitem{6751479}
{Fernando}, B., {Habrard}, A., {Sebban}, M., {Tuytelaars}, T.: Unsupervised
  visual domain adaptation using subspace alignment. In: 2013 IEEE ICCV (2013)

\bibitem{ganin2016domainadversarial}
Ganin, Y., Ustinova, E., Ajakan, H., Germain, P., Larochelle, H., Laviolette,
  F., Marchand, M., Lempitsky, V.: Domain-adversarial training of neural
  networks. J. Mach. Learn. Res.  (2016)

\bibitem{GottschalkD19}
Gottschalk, S., Demidova, E.: {EventKG} - the hub of event knowledge on the web
  - and biographical timeline generation. Semantic Web  \textbf{10}(6),
  1039--1070 (2019)

\bibitem{abs-2003-05370}
Jim{\'{e}}nez{-}Ruiz, E., Agibetov, A., Chen, J., Samwald, M., Cross, V.:
  Dividing the ontology alignment task with semantic embeddings and logic-based
  modules. In: {ECAI} 2020 (2020)

\bibitem{lample2018word}
Lample, G., Conneau, A., Ranzato, M., Denoyer, L., Jégou, H.: Word translation
  without parallel data. In: ICLR 2018 (2018)

\bibitem{madhavan2001generic}
Madhavan, J., Bernstein, P., Rahm, E.: Generic schema matching with cupid.
  Tech. rep. (August 2001)

\bibitem{994702}
{Melnik}, S., {Garcia-Molina}, H., {Rahm}, E.: Similarity flooding: a versatile
  graph matching algorithm and its application to schema matching. In: ICDE
  2002 (2002)

\bibitem{osmstats}
Neis, P.: {OSMstats}. \url{https://osmstats.neis-one.org/}, [Online; accessed
  10-April-2021]

\bibitem{DBLP:journals/semweb/NentwigHNR17}
Nentwig, M., Hartung, M., Ngomo, A.N., Rahm, E.: A survey of current link
  discovery frameworks. Semantic Web  \textbf{8}(3),  419--436 (2017)

\bibitem{NgoBT13}
Ngo, D., Bellahsene, Z., Todorov, K.: Opening the black box of ontology
  matching. In: Proc. of the {ESWC} 2013 (2013)

\bibitem{Ngomo:2011:LTA:2283696.2283783}
Ngomo, A.N., Auer, S.: {LIMES} - {A} time-efficient approach for large-scale
  link discovery on the web of data. In: {IJCAI} 2011. pp. 2312--2317 (2011)

\bibitem{Nkisi-OrjiWMHH18}
Nkisi{-}Orji, I., Wiratunga, N., Massie, S., Hui, K., Heaven, R.: Ontology
  alignment based on word embedding and random forest classification. In:
  {ECML} {PKDD} 2018

\bibitem{Otero-CerdeiraRG15}
Otero{-}Cerdeira, L., Rodr{\'{\i}}guez{-}Mart{\'{\i}}nez, F.J.,
  G{\'{o}}mez{-}Rodr{\'{\i}}guez, A.: Ontology matching: {A} literature review.
  Expert Syst. Appl.  \textbf{42}(2),  949--971 (2015)

\bibitem{DBLP:conf/semweb/PaulheimB13}
Paulheim, H., Bizer, C.: Type inference on noisy {RDF} data. In: {ISWC} 2013
  (2013)

\bibitem{10.1007/978-3-030-49461-2_34}
Pellissier~Tanon, T., Weikum, G., Suchanek, F.: {YAGO 4}: A reason-able
  knowledge base. In: ESWC 2020. Springer (2020)

\bibitem{DBLP:journals/vldb/RahmB01}
Rahm, E., Bernstein, P.A.: A survey of approaches to automatic schema matching.
  {VLDB} J.  \textbf{10}(4),  334--350 (2001)

\bibitem{10.1007/978-3-319-58068-5_7}
Sherif, M.A., Ngonga~Ngomo, A., Lehmann, J.: Wombat - {A} generalization
  approach for automatic link discovery. LNCS, vol. 10249, pp. 103--119.
  Springer (2017)

\bibitem{10.5555/2590208.2590210}
Stadler, C., Lehmann, J., H\"{o}ffner, K., Auer, S.: Linkedgeodata: A core for
  a web of spatial open data. Semant. Web  \textbf{3}(4),  333–354 (Oct 2012)

\bibitem{TEMPELMEIER2021349}
Tempelmeier, N., Demidova, E.: Linking openstreetmap with knowledge graphs -
  link discovery for schema-agnostic volunteered geographic information. Future
  Gener. Comput. Syst.  \textbf{116},  349--364 (2021)

\bibitem{inproceedingsunal}
Unal, O., Afsarmanesh, H.: Using linguistic techniques for schema matching. pp.
  115--120 (01 2006)

\bibitem{www2009227}
Volz, J., Bizer, C., Gaedke, M., Kobilarov, G.: Silk - {A} link discovery
  framework for the web of data. In: Proceedings of the {WWW2009} Workshop on
  Linked Data on the Web, {LDOW} 2009. vol.~538

\bibitem{articlewikidata}
Vrandečić, D., Krötzsch, M.: Wikidata: A free collaborative knowledgebase.
  Communications of the ACM  \textbf{57},  78--85 (09 2014)

\bibitem{xiang-etal-2015-ersom}
Xiang, C., Jiang, T., Chang, B., Sui, Z.: {ERSOM}: A structural ontology
  matching approach using automatically learned entity representation. In:
  EMNLP (2015)

\bibitem{ZhangB20}
Zhang, S., Balog, K.: Web table extraction, retrieval, and augmentation: {A}
  survey. {ACM} Trans. Intell. Syst. Technol.  \textbf{11}(2),  13:1--13:35
  (2020)

\end{thebibliography}
